\documentclass{article}
\usepackage{arxiv}
\usepackage[utf8]{inputenc}
\usepackage[T1]{fontenc}
\usepackage{hyperref}
\usepackage{url}
\usepackage{booktabs}
\usepackage{amsfonts}
\usepackage{nicefrac}
\usepackage{subcaption}
\usepackage{microtype}
\usepackage{graphicx}

\usepackage[backend=biber,style=numeric,sorting=none,maxnames=3,minnames=1]{biblatex}
\title{Performance Trade-offs of Optimizing Small Language Models for E-Commerce}

\author{
 Josip Tomo Licardo \\
  Faculty of Informatics\\
  Juraj Dobrila University of Pula\\
  Zagrebačka 30\\
  52100 Pula, Croatia \\
  \texttt{jlicardo@unipu.hr} \\
   \And
 Nikola Tanković \\
  Faculty of Informatics\\
  Juraj Dobrila University of Pula\\
  Zagrebačka 30\\
  52100 Pula, Croatia \\
  \texttt{ntankov@unipu.hr} \\
}

\begin{document}
\maketitle
\begin{abstract}
Large Language Models (LLMs) offer state-of-the-art performance in natural language understanding and generation tasks. However, the deployment of leading commercial models for specialized tasks, such as e-commerce, is often hindered by high computational costs, latency, and operational expenses. This paper investigates the viability of smaller, open-weight models as a resource-efficient alternative. We present a methodology for optimizing a one-billion-parameter Llama 3.2 model for multilingual e-commerce intent recognition. The model was fine-tuned using Quantized Low-Rank Adaptation (QLoRA) on a synthetically generated dataset designed to mimic real-world user queries. Subsequently, we applied post-training quantization techniques, creating GPU-optimized (GPTQ) and CPU-optimized (GGUF) versions. Our results demonstrate that the specialized 1B model achieves 99\% accuracy, matching the performance of the significantly larger GPT-4.1 model. A detailed performance analysis revealed critical, hardware-dependent trade-offs: while 4-bit GPTQ reduced VRAM usage by 41\%, it paradoxically slowed inference by 82\% on an older GPU architecture (NVIDIA T4) due to dequantization overhead. Conversely, GGUF formats on a CPU achieved a speedup of up to 18× in inference throughput and a reduction of over 90\% in RAM consumption compared to the FP16 baseline. We conclude that small, properly optimized open-weight models are not just a viable but a more suitable alternative for domain-specific applications, offering state-of-the-art accuracy at a fraction of the computational cost.
\end{abstract}

\keywords{Large Language Models \and E-commerce \and Fine-tuning \and QLoRA \and Quantization \and GPTQ}

\section{Introduction}

The field of artificial intelligence has been fundamentally reshaped by the advent of Large Language Models (LLMs), sophisticated deep learning systems that demonstrate a remarkable capacity to understand, generate, and reason with human language \cite{vaswaniAttentionAllYou2023a}. Foundational models such as Meta's Llama 3 series \cite{grattafioriLlama3Herd2024b}, Google's Gemma 3 \cite{teamGemma3Technical2025a}, and Alibaba's Qwen3 \cite{yangQwen3TechnicalReport2025a} now represent the state-of-the-art, powering applications that span from complex code generation to nuanced creative writing and scientific discovery. Their advanced capabilities have unlocked new paradigms for human-computer interaction and process automation across countless industries.

One of the most promising domains for LLM application is e-commerce, a highly competitive landscape where user experience is a critical determinant of success. The ability of a system to accurately interpret and act upon user requests expressed in natural, often informal, language is a significant competitive advantage. This paradigm, known as conversational commerce, aims to create more intuitive and personalized shopping experiences \cite{sidlauskieneAIbasedChatbotsConversational2023}. Tasks such as robust product classification from noisy user descriptions \cite{gholamianLLMBasedRobustProduct2024} and understanding complex user intent are central to this vision \cite{zhouUsagecentricTakeIntent2024}. Major industry players like eBay have already begun developing their own in-house, domain-specific LLMs to capitalize on these opportunities, underscoring the strategic importance of this technology \cite{heroldLiLiuMEBaysLarge2024}.

However, the deployment of the most powerful, state-of-the-art commercial models, such as OpenAI's GPT-4, presents significant practical and economic barriers. These models are computationally massive, and their use via API services generates continuous operational costs, creates vendor lock-in, and raises concerns regarding data privacy and security. Furthermore, the immense energy, water, and carbon footprint associated with serving billions of queries from these large-scale models is a growing concern for sustainable AI development \cite{jeghamHowHungryAI2025, fernandezEnergyConsiderationsLarge2025, husomPricePromptingProfiling2024}. The high cost and resource intensity of these "frontier" models thus create a substantial obstacle to their widespread adoption, particularly for small to medium-sized enterprises.

In response to these challenges, a powerful trend has emerged: the rise of smaller, highly optimized, open-weight language models. A growing body of research demonstrates that these smaller models, when specialized for a specific domain, can achieve performance comparable or even superior to that of much larger, general-purpose models. For instance, fine-tuned small models have been shown to outperform GPT-4 on specific tasks like arithmetic \cite{liuGoatFinetunedLLaMA2023a}, mental health understanding \cite{jiaScaleSmallLanguage2025}, and developing pedagogical tools \cite{solanoNarrowingGapSupervised2025}. A large-scale study fine-tuning over 300 models confirmed that specialized models can consistently rival GPT-4 on narrow tasks, suggesting that a "many small models" approach is a viable and efficient strategy \cite{zhaoLoRALand3102024}. This paradigm shift from monolithic, "one-size-fits-all" models to a diverse ecosystem of specialized agents motivates our research.

Two key technologies are central to unlocking the potential of these smaller models: parameter-efficient fine-tuning (PEFT) and post-training quantization (PTQ). PEFT techniques allow a pretrained model to be adapted to a new task by training only a small fraction of its parameters. Low-Rank Adaptation (LoRA) has become a standard approach \cite{huLoRALowRankAdaptation2021a}, and its successor, Quantized LoRA (QLoRA), further democratized this process by enabling the fine-tuning of large models on consumer-grade hardware through aggressive 4-bit quantization of the base model during training \cite{dettmersQLoRAEfficientFinetuning2023a}.

Concurrently, post-training quantization has become essential for efficient inference. PTQ methods reduce the numerical precision of a model's weights (and sometimes activations) after training, leading to significant reductions in memory footprint and potential increases in inference speed. Techniques like GPTQ \cite{frantarGPTQAccuratePostTraining2023a} and AWQ \cite{linAWQActivationawareWeight2024a} have demonstrated the ability to compress models to 4-bit precision with minimal accuracy loss. However, recent research highlights that the benefits of quantization are not universal; they are highly dependent on the model size, task, and underlying hardware architecture, creating a complex web of trade-offs between performance, energy, and output quality \cite{leeExploringTradeOffsQuantization2025a, shiSystematicCharacterizationLLM2025, zhaoBenchmarkingPostTrainingQuantization2025}.

This paper bridges these research threads by conducting a comprehensive, hardware-aware investigation into the practical viability of a small, optimized open-weight model for a critical e-commerce task. We hypothesize that by combining QLoRA-based fine-tuning with aggressive post-training quantization, it is possible to specialize a 1 billion parameter model to achieve accuracy parity with the much larger, state-of-the-art GPT-4.1 on a structured intent recognition task. Crucially, we extend beyond mere accuracy to analyze the real-world performance trade-offs, measuring inference speed, memory consumption, and energy efficiency on both GPU and CPU environments to provide a holistic view of the model's operational characteristics.

Our contributions are fourfold. First, we introduce an end-to-end methodology for building a highly efficient, specialized language model for multilingual e-commerce intent recognition, including a novel synthetic data generation process based on "metaprompting". Second, we supply empirical evidence that a properly fine-tuned 1-billion-parameter Llama 3.2 model reaches 99\% accuracy on this task, matching GPT-4.1. Third, we present a detailed, hardware-aware performance analysis that surfaces nuanced, sometimes counter-intuitive effects of quantization, showing that 4-bit GPTQ can slow inference on older GPUs while GGUF formats yield significant speedups on CPUs. Fourth, we release our multilingual synthetic dataset for e-commerce intent recognition to support further research and reproducibility.

The remainder of this paper is organized as follows. Section 2 discusses the background and related work in model specialization and optimization. Section 3 details our methodology, including task definition, dataset generation, and the experimental pipeline. Section 4 presents our results on model accuracy and operational performance. Section 5 discusses the implications of our findings, and Section 6 concludes the paper with a summary and directions for future work.

\section{Background and Related Work}

The rapid evolution of LLMs is characterized by a dual trend: the scaling of massive, general-purpose models and the proliferation of smaller, highly specialized models. While large models set performance benchmarks, their practical deployment is often limited. This has spurred significant research into methodologies for creating and optimizing smaller models that are both powerful and efficient. Our work is situated at the intersection of three key research areas: domain specialization through fine-tuning, performance optimization via quantization, and the generation of high-quality data for these processes.

\subsection{From Generalists to Specialists: The Power of Fine-Tuning}
The prevailing paradigm is shifting from relying on a single, monolithic model to deploying smaller models tailored for specific tasks. A growing body of evidence shows that this specialization is highly effective. Large-scale studies, such as the one conducted in LoRA Land, have demonstrated that fine-tuning a diverse set of 7B models with LoRA can result in performance that consistently rivals or exceeds that of GPT-4 on many narrow tasks \cite{zhaoLoRALand3102024}. This effect holds across various domains; for instance, the Goat model, a fine-tuned LLaMA, surpassed GPT-4 on complex arithmetic tasks \cite{liuGoatFinetunedLLaMA2023a}, and specialized models have shown comparable performance to GPT-4 in sensitive fields such as mental health understanding \cite{jiaScaleSmallLanguage2025} and for creating pedagogical tools \cite{solanoNarrowingGapSupervised2025}. This trend is also prominent in industry, with companies such as eBay developing their own in-house, e-commerce-focused LLMs to gain a competitive edge \cite{heroldLiLiuMEBaysLarge2024}.

The primary mechanism for achieving this specialization is fine-tuning. However, full fine-tuning, which updates all of a model's billions of parameters, is resource-prohibitive. This has led to the dominance of Parameter-Efficient Fine-Tuning (PEFT) methods. Low-Rank Adaptation (LoRA) is a foundational PEFT technique that freezes the pretrained model weights and injects small, trainable low-rank matrices, dramatically reducing the number of trainable parameters and memory requirements \cite{huLoRALowRankAdaptation2021a}. Building on this, Quantized LoRA (QLoRA) introduced a breakthrough by quantizing the base model to an aggressive 4-bit precision during fine-tuning. This, combined with innovations such as the NormalFloat4 (NF4) data type and paged optimizers, enables the fine-tuning of very large models (e.g., 65B) on a single consumer GPU while maintaining the performance of 16-bit fine-tuning \cite{dettmersQLoRAEfficientFinetuning2023a}. The PEFT landscape continues to evolve with more advanced techniques such as DoRA, which dynamically allocates rank during training \cite{maoDoRAEnhancingParameterEfficient2024}, further enhancing the toolkit for creating specialized models.

\subsection{Post-Training Quantization for Efficient Inference}
While QLoRA uses quantization during training, Post-Training Quantization (PTQ) is a critical step for optimizing models for efficient deployment. The primary goals of PTQ are to reduce the model's memory footprint, decrease inference latency, and lower energy consumption \cite{fernandezEnergyConsiderationsLarge2025, maliakelInvestigatingEnergyEfficiency2025}. Numerous PTQ methods have been developed, with GPTQ and AWQ being among the most prominent. GPTQ is a one-shot weight quantization method that uses approximate second-order information to iteratively quantize weights while compensating for errors, achieving high accuracy even at 3 or 4-bit precision \cite{frantarGPTQAccuratePostTraining2023a}. In contrast, Activation-aware Weight Quantization (AWQ) identifies that a small fraction of weights are disproportionately important for performance. It protects these salient weights by scaling them up before uniform quantization, a simple yet highly effective strategy \cite{linAWQActivationawareWeight2024a}. Other methods such as SmoothQuant tackle the problem by mathematically migrating quantization difficulty from activations, which often have problematic outliers, to the more easily quantized weights \cite{xiaoSmoothQuantAccurateEfficient2024a}.

However, the benefits of quantization are not a "free lunch." Recent comprehensive benchmarks reveal a complex landscape of trade-offs. The effectiveness of a given PTQ method is highly dependent on the model size, the target task, and the underlying hardware \cite{leeExploringTradeOffsQuantization2025a, leeComprehensiveEvaluationQuantized2024, zhaoBenchmarkingPostTrainingQuantization2025}. For instance, smaller models often suffer greater accuracy degradation from 4-bit quantization than larger ones \cite{leeExploringTradeOffsQuantization2025a}. Furthermore, quantization can disproportionately affect certain capabilities, such as mathematical reasoning \cite{liQuantizationMeetsReasoning2025}, and may even alter a model's truthfulness under certain conditions \cite{fuQuantizedDeceptiveMultiDimensional2025}. A systematic characterization of LLM quantization shows that choices regarding tensor parallelism and GPU architecture can dramatically alter the realized gains in latency and energy, underscoring the need for hardware-aware evaluation \cite{shiSystematicCharacterizationLLM2025}.

\subsection{Synthetic Data and Structured Output Generation}
The success of fine-tuning is fundamentally dependent on the quality and volume of the training data. When domain-specific data is scarce, synthetic data generation using LLMs has become an indispensable technique \cite{nadasSyntheticDataGeneration2025a}. The Self-Instruct methodology first demonstrated that an LLM could bootstrap its own instruction-following dataset from a small seed set \cite{wangSelfInstructAligningLanguage2023}. This concept has evolved into sophisticated, agentic pipelines such as AgentInstruct \cite{mitraAgentInstructGenerativeTeaching2024} and MetaSynth \cite{riazMetaSynthMetaPromptingDrivenAgentic2025a}, which use multi-LLM systems to generate, refine, and diversify synthetic data at a massive scale. The quality of the data generator is key, as different LLMs exhibit complementary strengths in this role \cite{kimEvaluatingLanguageModels2025}.

A specific challenge within this domain, and one central to our work, is generating structured outputs such as JSON. This capability is critical for integrating LLMs into automated workflows and RAG systems. Recent research has focused on benchmarking this capability \cite{yangStructEvalBenchmarkingLLMs2025a, shortenStructuredRAGJSONResponse2024} and developing techniques to enforce strict schema adherence, such as reinforcement learning strategies \cite{agarwalThinkJSONReinforcement2025}. Studies comparing fine-tuned small models against prompted large models for generating structured low-code workflows have found that targeted fine-tuning often yields more robust and domain-aware results \cite{ayalaFineTuneSLMPrompt2025}. However, it is also known that imposing strict format restrictions can sometimes impair a model's underlying reasoning ability, suggesting a trade-off between structural rigidity and performance \cite{tamLetMeSpeak2024}. Our work builds on these insights by using an advanced LLM to generate a structured, noisy, multilingual dataset tailored for a real-world e-commerce task, providing the foundation for our specialization experiments.

\section{Methodology}

To systematically evaluate the performance trade-offs of optimizing small language models, we designed a multi-stage experimental pipeline. The process encompasses task definition, synthetic dataset generation, model selection, fine-tuning, post-training quantization, and a dual-pronged evaluation of both accuracy and operational performance.

\subsection{Task Definition and Dataset Generation}
The core task of our experiment is structured intent extraction from natural language user queries in an e-commerce context. The model's objective is to parse a user's free-form text request for managing a shopping cart and output a structured JSON object containing three key fields: `action` (e.g., "add" or "remove"), `product` (the canonical name of the item), and `quantity` (an integer).

Given the lack of publicly available, multilingual datasets for this specific task, we generated a high-quality synthetic dataset. We employed a "metaprompting" strategy, using the GPT-4.1 model as a sophisticated data generator. A Python script systematically constructed detailed prompts that guided the generator model to produce diverse and realistic examples.  

For each example, the script programmatically defined the ground-truth parameters: the language (cycling through English, Croatian, and Spanish), the action (either `add` or `remove`), a product randomly selected from a predefined catalog, and a quantity chosen via a weighted random selection to mimic realistic purchasing patterns. To introduce stylistic diversity, a wide range of prompt templates was employed, generating user expressions that varied from formal requests to brief imperative commands.  

To further enhance realism and robustness, the script strategically injected noise. This included linguistic noise such as typos or slang like “pls” and “thx”, contextual noise such as greetings, emojis, or unrelated brand names, and instances of code-switching, for example embedding English phrases like “free shipping” into Croatian sentences.  

This pipeline produced a dataset of 3,000 examples. An example of a data point is shown below:
\begin{verbatim}
User input: "Can you delet 12 lip balms for me?"
->
JSON: {"action": "remove", "product": "Lip Balm", "quantity": 12}
\end{verbatim}
The dataset, named \texttt{jtlicardo/ecommerce-intent-3k}, was published on the Hugging Face Hub \footnote{\url{https://huggingface.co/datasets/jtlicardo/ecommerce-intent-3k}} and split into a 90\% training set and a 10\% validation set.

\subsection{Models and Baselines}
Our investigation centered on optimizing a small, open-weight model. The primary model for fine-tuning and quantization was \textbf{Llama 3.2 1B}, a recent and efficient model from Meta \cite{grattafioriLlama3Herd2024b}. To contextualize its performance, we also evaluated a representative set of other open-weight models, including \textbf{Gemma 3 1B} and \textbf{Gemma 2 2B} from Google \cite{teamGemma3Technical2025a}, and \textbf{Qwen 2.5 1.5B} and \textbf{Qwen 2.5 3B} from Alibaba \cite{yangQwen3TechnicalReport2025a}.

To establish a state-of-the-art performance benchmark, we used leading commercial models from OpenAI's \textbf{GPT-4.1 series}. These proprietary models serve as an upper bound for accuracy on the given task and were evaluated using a few-shot prompting approach, where several examples were provided in the prompt to guide the model's response format.

\subsection{Experimental Pipeline}
The core of our experiment involved a two-stage process: specializing the base model for our task via fine-tuning, and then optimizing the specialized model for efficient deployment via quantization.

\subsubsection{Fine-tuning}
We employed \textbf{QLoRA (Quantized Low-Rank Adaptation)} \cite{dettmersQLoRAEfficientFinetuning2023a}, a parameter-efficient fine-tuning technique, to train our model. The process was orchestrated using the `transformers`, `peft`, and `bitsandbytes` libraries. The base Llama 3.2 1B model was loaded with its weights quantized to 4-bit precision using the NormalFloat4 (NF4) data type. A LoRA adapter was then applied with the following key hyperparameters: rank (\texttt{r}) of 8, alpha (\texttt{lora\_alpha}) of 16, and a dropout rate of 0.1. The adapter targeted all linear projections within the self-attention and feed-forward network blocks. The model was trained for 5 epochs using the `SFTTrainer` with a batch size of 8 and a learning rate of 2e-5. Crucially, the loss was calculated only on the completion part of the sequence (the JSON output), focusing the learning process exclusively on generating the correct structured data.

\subsubsection{Post-Training Quantization}
After fine-tuning, the trained LoRA adapter was merged with the original, full-precision (FP16) Llama 3.2 1B base model. This step created a single, unified, specialized model in FP16 format, which served as the foundation for all subsequent quantization.
\begin{enumerate}
    \item \textbf{GPTQ for GPU:} We used the 'auto-gptq' library to apply \textbf{GPTQ (Generative Pre-trained Transformer Quantization)} \cite{frantarGPTQAccuratePostTraining2023a}. Using a calibration set of 300 random samples from our training data, we quantized the merged FP16 model to 4-bit precision, creating a version optimized for GPU inference.
    \item \textbf{GGUF for CPU:} We used the 'llama.cpp' toolchain to convert the merged FP16 model into the GGUF format, which is highly optimized for CPU inference. We generated three distinct versions to analyze the impact of bit depth: an aggressive 3-bit version (Q3\_K\_M), a balanced 4-bit version (Q4\_K\_M), and a high-quality 5-bit version (Q5\_K\_M).
\end{enumerate}

\subsection{Evaluation Framework}
To ensure a rigorous and fair comparison, we used a separate, unseen test set of 100 examples, \texttt{jtlicardo/ecommerce-intent-eval} \footnote{\url{https://huggingface.co/datasets/jtlicardo/ecommerce-intent-eval}}, structured identically to the training data.

\subsubsection{Accuracy Metric}
Performance was measured using \textbf{Exact Match Accuracy}. For a prediction to be considered correct, the generated output had to be a syntactically valid JSON object, and all three key-value pairs (`action`, `product`, and `quantity`) had to be identical to the ground truth labels. This strict metric was chosen because partial correctness in this e-commerce task can lead to significant functional errors in a production system.

\subsubsection{Performance Metrics and Hardware}
Beyond accuracy, we profiled the key operational characteristics of our fine-tuned Llama 3.2 1B model in its FP16 and quantized forms. We measured inference speed, reported in tokens generated per second, memory consumption, expressed in gigabytes (GB) of VRAM for GPU tests and RAM for CPU tests, and energy consumption, where power draw in Watts was monitored using \texttt{nvidia-smi} during GPU tests to calculate energy efficiency. GPU-based evaluations (FP16 vs.~GPTQ) were conducted on an \textbf{NVIDIA T4} GPU within the Google Colab environment, while CPU-based evaluations (GGUF variants) were performed on a local machine equipped with an \textbf{AMD Ryzen 7 5800HS} processor.

\section{Results}

This section presents the empirical findings of our study. We first compare the accuracy of all evaluated models to establish a performance baseline and validate our primary hypothesis. We then provide a detailed analysis of the operational characteristics of our optimized model, examining the hardware-dependent trade-offs of quantization on both GPU and CPU platforms.

\subsection{Model Accuracy Comparison}

The primary goal of our experiment was to determine if a small, specialized open-weight model could achieve accuracy parity with large-scale commercial models. The exact match accuracy scores for all models on the unseen test set are presented in Table \ref{tab:accuracy_results}.

\begin{table}[htbp]
  \caption{Comparison of Exact Match Accuracy Across All Evaluated Models.}
  \label{tab:accuracy_results}
  \centering
  \begin{tabular}{@{}llc@{}}
    \toprule
    \textbf{Model Family} & \textbf{Model Variant} & \textbf{Accuracy} \\
    \midrule
    \addlinespace[0.3em]
    \multicolumn{3}{@{}l}{\textit{Commercial Baselines (Few-shot)}} \\
    \cmidrule(l){2-3}
    \addlinespace[0.3em]
    \raisebox{-0.4\normalbaselineskip}[0pt][0pt]{GPT (OpenAI)} & GPT 4.1 & 1.00 \\
     & GPT 4.1-mini & 1.00 \\
     & GPT 4.1-nano & 0.99 \\
    \midrule
    \addlinespace[0.3em]
    \multicolumn{3}{@{}l}{\textit{Open-Weight Models (Base and Fine-tuned)}} \\
    \cmidrule(l){2-3}
    \addlinespace[0.3em]
    \raisebox{-1.0\normalbaselineskip}[0pt][0pt]{Qwen (Alibaba)} & Qwen 2.5 1.5B & 0.86 \\
     & Qwen 2.5 1.5B Instruct & 0.87 \\
     & Qwen 2.5 3B & 0.96 \\
     & Qwen 2.5 3B Instruct & 0.96 \\
    \addlinespace[0.3em]
    \cmidrule(l){2-3}
    \addlinespace[0.3em]
    \raisebox{-0.7\normalbaselineskip}[0pt][0pt]{Gemma (Google)} & Gemma 3 1B & 0.78 \\
     & Gemma 3 1B (finetune) & 0.90 \\
     & Gemma 2 2B & 0.86 \\
    \addlinespace[0.3em]
    \cmidrule(l){2-3}
    \addlinespace[0.3em]
    \raisebox{-1.5\normalbaselineskip}[0pt][0pt]{Llama (Meta)} & Llama 3.2 1B & 0.82 \\
     & Llama 3.2 1B (finetune) & \textbf{0.99} \\
     & Llama 3.2 1B (finetune, GPTQ 4-bit) & \textbf{0.99} \\
     & Llama 3.2 1B (finetune, GGUF 3-bit) & 0.60 \\
     & Llama 3.2 1B (finetune, GGUF 4-bit) & 0.89 \\
     & Llama 3.2 1B (finetune, GGUF 5-bit) & \textbf{0.99} \\
    \bottomrule
  \end{tabular}
\end{table}

\begin{figure}[htbp]
  \centering
  \includegraphics[width=0.8\textwidth]{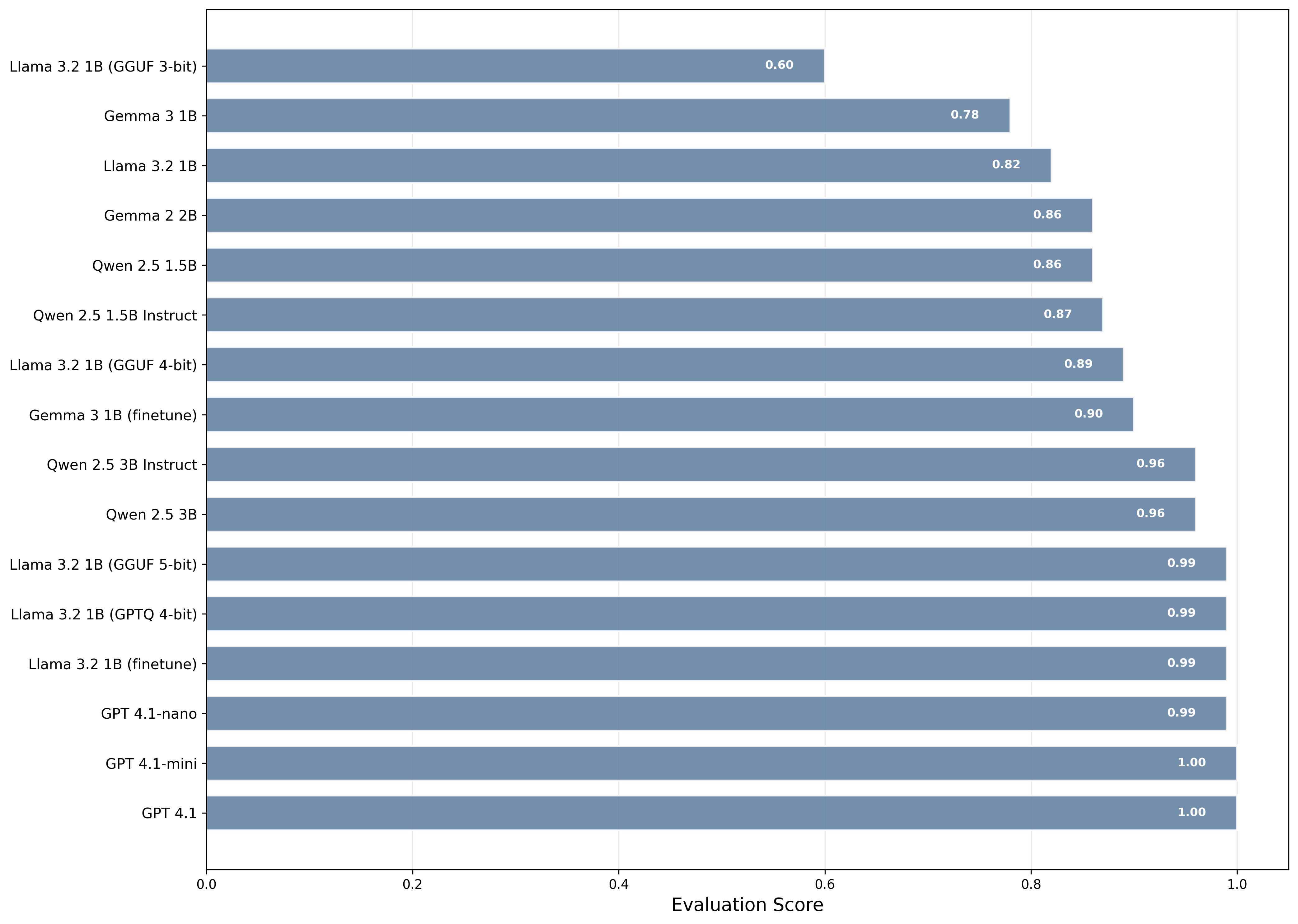}
  \caption{Overall Model Performance Comparison. The fine-tuned Llama 3.2 1B model and its high-quality quantized variants achieve accuracy scores comparable to the top commercial baselines.}
  \label{fig:model_performance_comparison}
\end{figure}

As expected, the commercial GPT-4.1 series models set a high bar, achieving near-perfect scores. The base open-weight models showed varied but generally respectable performance, with the Llama 3.2 1B base model achieving an accuracy of 0.82. The most significant finding is the dramatic performance increase after fine-tuning. The Llama 3.2 1B model, after fine-tuning with QLoRA, saw its accuracy jump to 0.99, an improvement of 21\%. This result places the specialized 1B parameter model on equal footing with the much larger GPT-4.1-nano and effectively matches the performance of the top-tier commercial models. This outcome provides strong support for our central hypothesis. The overall performance landscape is visualized in Figure \ref{fig:model_performance_comparison}.

\subsection{The Impact of Quantization on Accuracy}
A key aspect of our investigation was to determine how post-training quantization affects the accuracy of the specialized model. The results, shown in Table \ref{tab:accuracy_results}, reveal that modern quantization techniques can preserve performance with high fidelity. Both the 4-bit GPTQ version and the 5-bit GGUF (Q5\_K\_M) version fully retained the 0.99 accuracy of the FP16 fine-tuned model.

However, more aggressive quantization introduces a clear trade-off. The 4-bit GGUF (Q4\_K\_M) version experienced a noticeable drop in accuracy to 0.89. The most aggressive 3-bit GGUF (Q3\_K\_M) variant suffered a catastrophic performance collapse, with accuracy plummeting to 0.60. This demonstrates the existence of a "quantization cliff," \cite{ahmadianIntriguingPropertiesQuantization2023} a threshold below which the model loses the numerical precision required to execute its specialized task effectively.

\subsection{GPU Performance Profile (FP16 vs. GPTQ)}
We profiled the FP16 fine-tuned model against its 4-bit GPTQ version on an NVIDIA T4 GPU to assess performance in a typical accelerated environment. The results reveal a nuanced and counterintuitive relationship between quantization, memory, and speed on this specific hardware architecture.

\begin{figure}[htbp]
  \centering
  \includegraphics[width=0.8\textwidth]{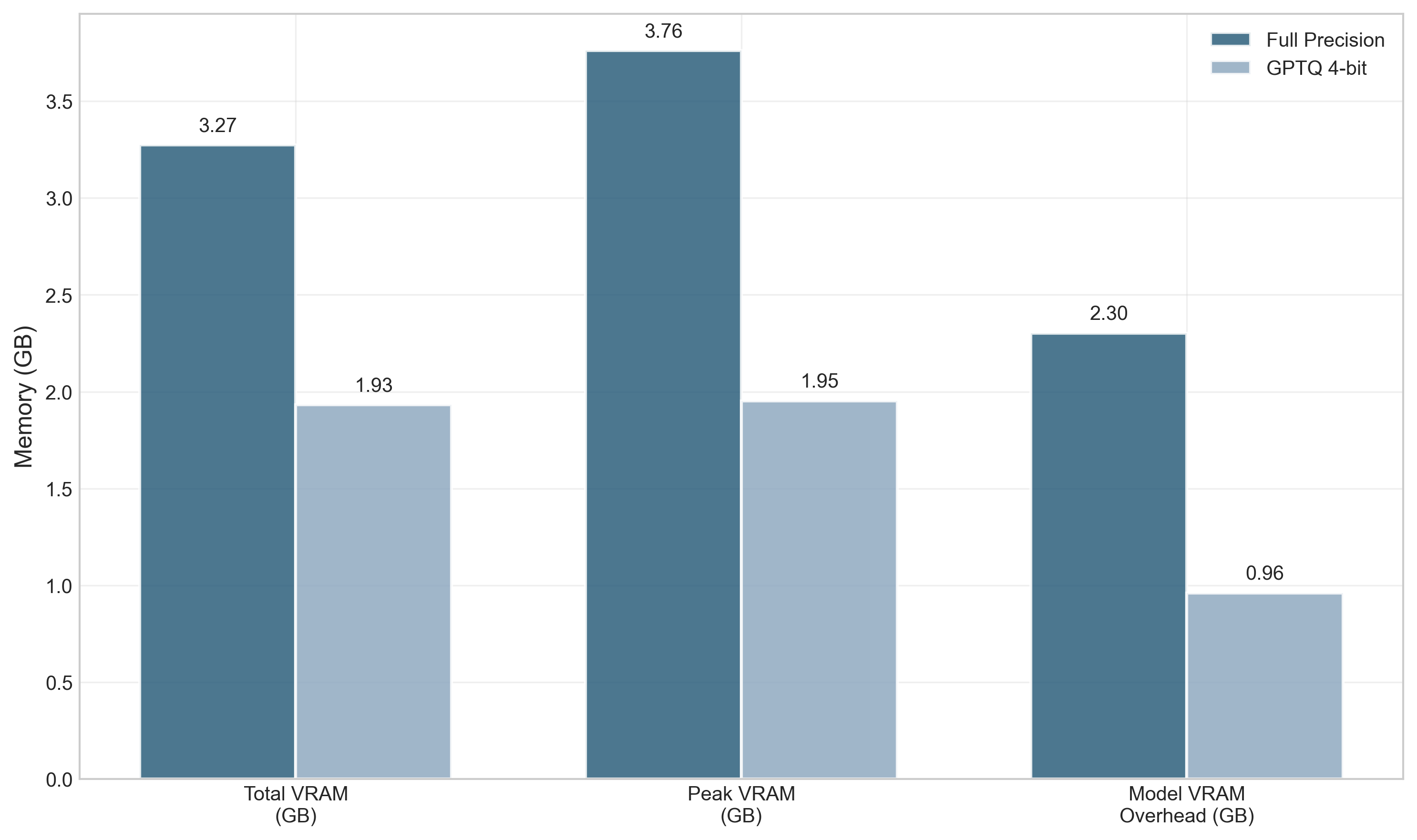}
  \caption{GPU Memory Comparison. GPTQ 4-bit quantization significantly reduces total VRAM, peak VRAM, and the memory required for model parameters alone.}
  \label{fig:gpu_memory}
\end{figure}

As shown in Figure \ref{fig:gpu_memory}, quantization delivered substantial memory savings. The 4-bit GPTQ model reduced total VRAM consumption from 3.27 GB to 1.93 GB (a 41\% reduction) and cut the model's parameter-only footprint from 2.30 GB to just 0.96 GB.

\begin{figure}[htbp]
  \centering
  \begin{subfigure}[b]{0.48\textwidth}
    \includegraphics[width=0.9\textwidth]{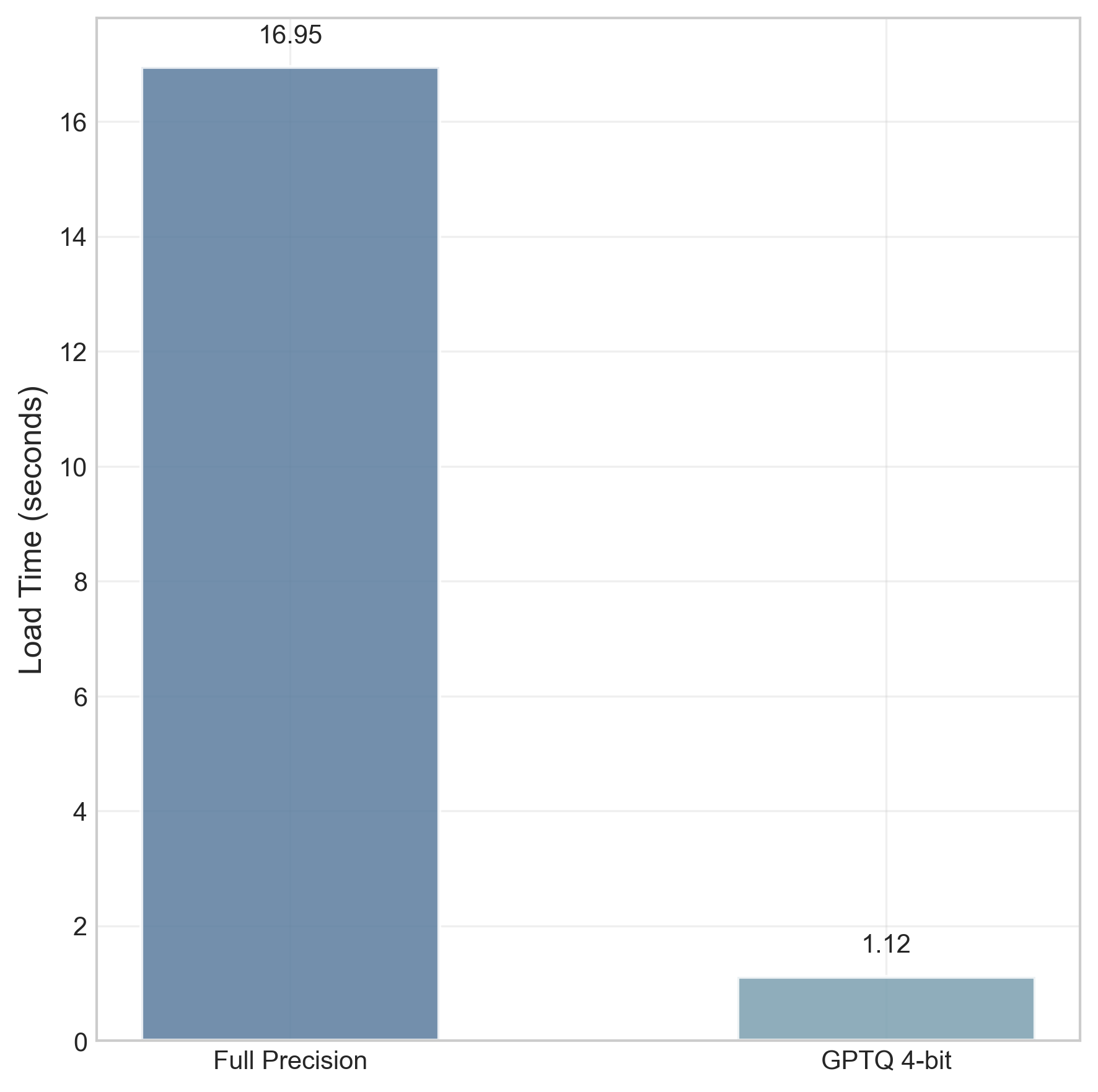}
    \caption{Load time (s).}
    \label{fig:gpu_load_time}
  \end{subfigure}\hfill
  \begin{subfigure}[b]{0.48\textwidth}
    \includegraphics[width=0.9\textwidth]{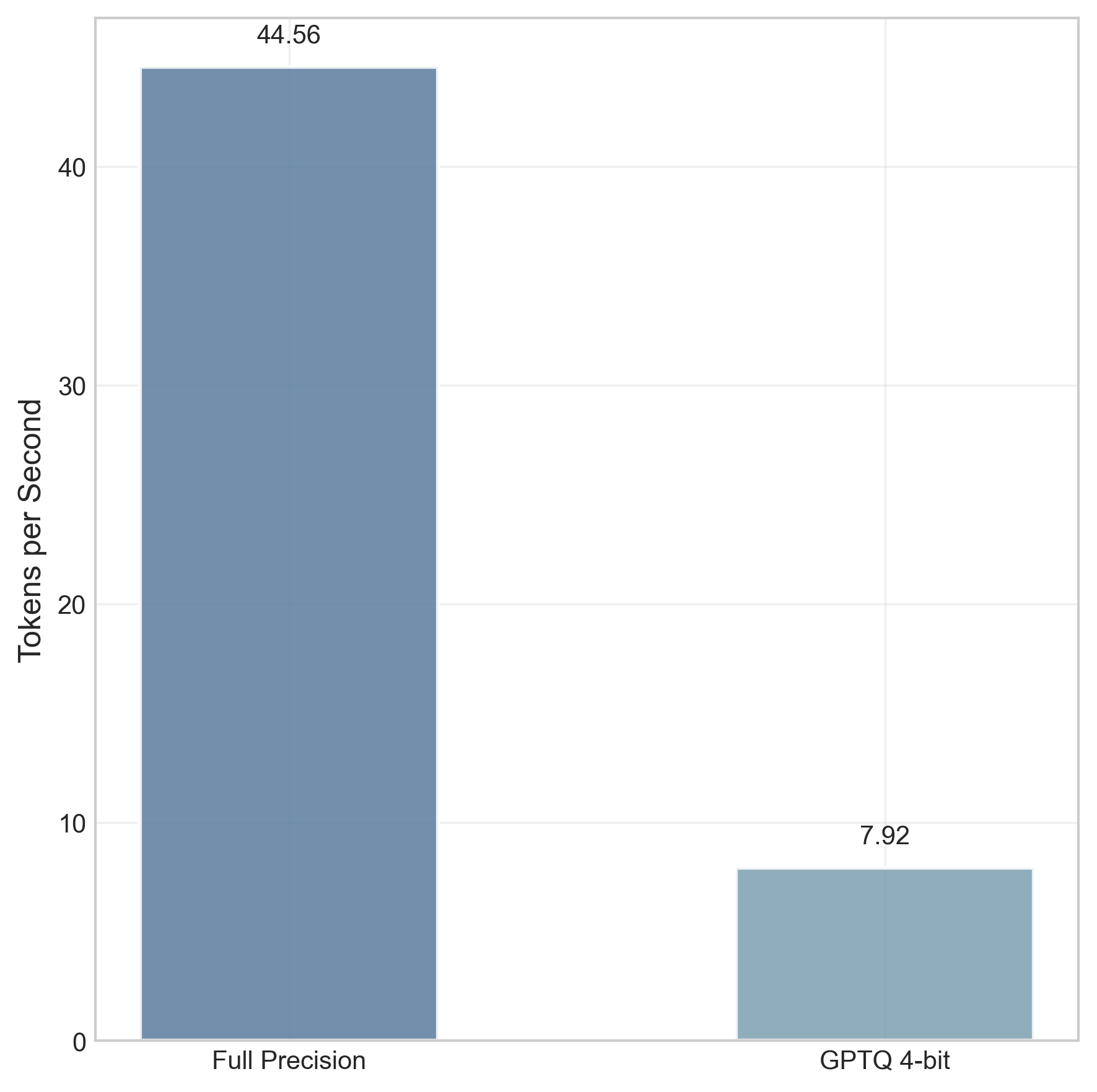}
    \caption{Tokens per second.}
    \label{fig:gpu_tokens}
  \end{subfigure}
  \caption{GPU speed profile on NVIDIA T4. GPTQ loads much faster but runs with lower throughput.}
  \label{fig:gpu_speed}
\end{figure}

\begin{figure}[htbp]
  \centering
  \includegraphics[width=0.5\textwidth]{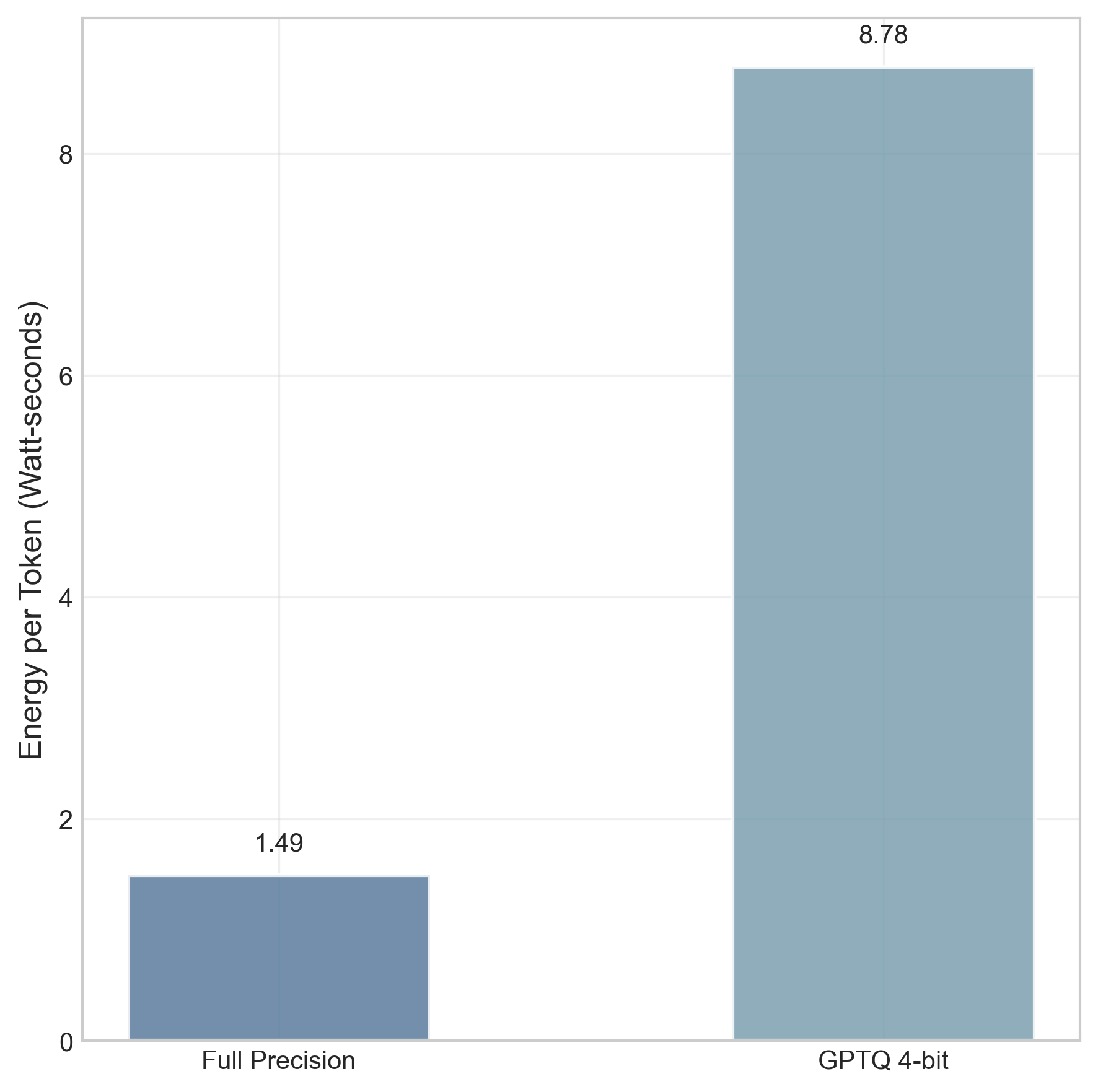}
  \caption{Energy per token on NVIDIA T4. GPTQ consumes substantially more energy per generated token.}
  \label{fig:gpu_energy}
\end{figure}

Despite these memory benefits, the GPTQ model was significantly slower during inference, as detailed in Figures~\ref{fig:gpu_speed} and~\ref{fig:gpu_energy}. While the model load time decreased dramatically by 93.4\% (from 16.95\,s to 1.12\,s), the inference speed dropped from 44.56 tokens/second for the FP16 model to just 7.92 tokens/second for the GPTQ version, an 82.2\% slowdown. The slowdown suggests that the T4 GPU did not execute computations directly in 4-bit. Instead, it appears that the quantized weights were converted back to higher precision during inference, adding overhead. Consequently, the energy consumed per token was 489.3\% higher for the quantized model, making it less efficient for inference on this architecture.

\subsection{CPU Performance Profile (GGUF)}
In stark contrast to the GPU results, the GGUF-quantized models demonstrated exceptional performance gains on a CPU (AMD Ryzen 7 5800HS), leveraging the highly optimized `llama.cpp` library.

\begin{figure}[htbp]
  \centering
  \includegraphics[width=0.7\textwidth]{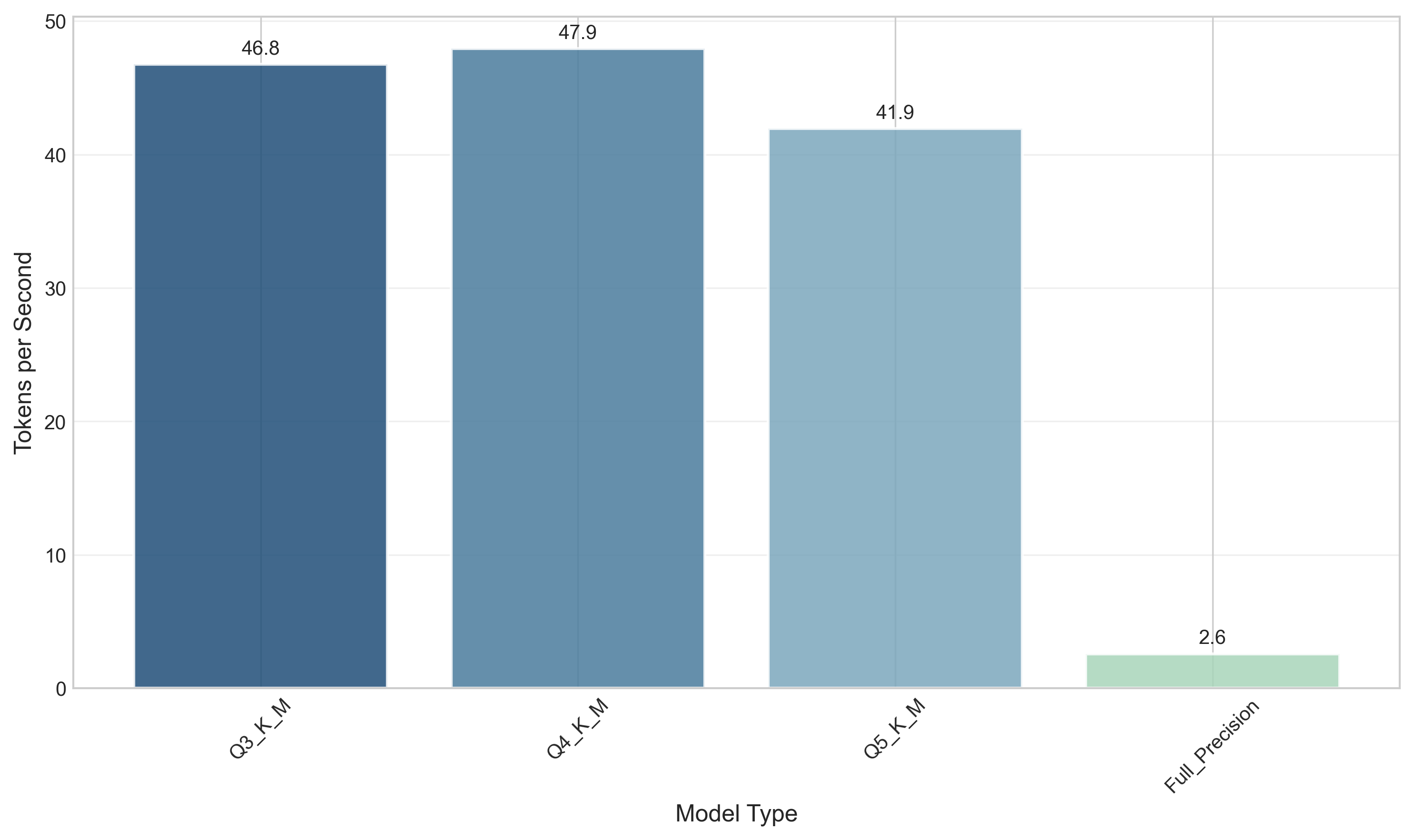}
  \caption{CPU Inference Performance. All GGUF-quantized models offer a dramatic speedup over the full-precision baseline.}
  \label{fig:cpu_speed}
\end{figure}

As shown in Figure \ref{fig:cpu_speed}, the FP16 model was impractically slow, achieving only 2.6 tokens/second. In contrast, all quantized GGUF versions provided a massive speedup. The 4-bit (Q4\_K\_M) version was the fastest, peaking at 47.9 tokens/second, representing an 18× improvement compared to the FP16 baseline.

\begin{figure}[htbp]
  \centering
  \includegraphics[width=\textwidth]{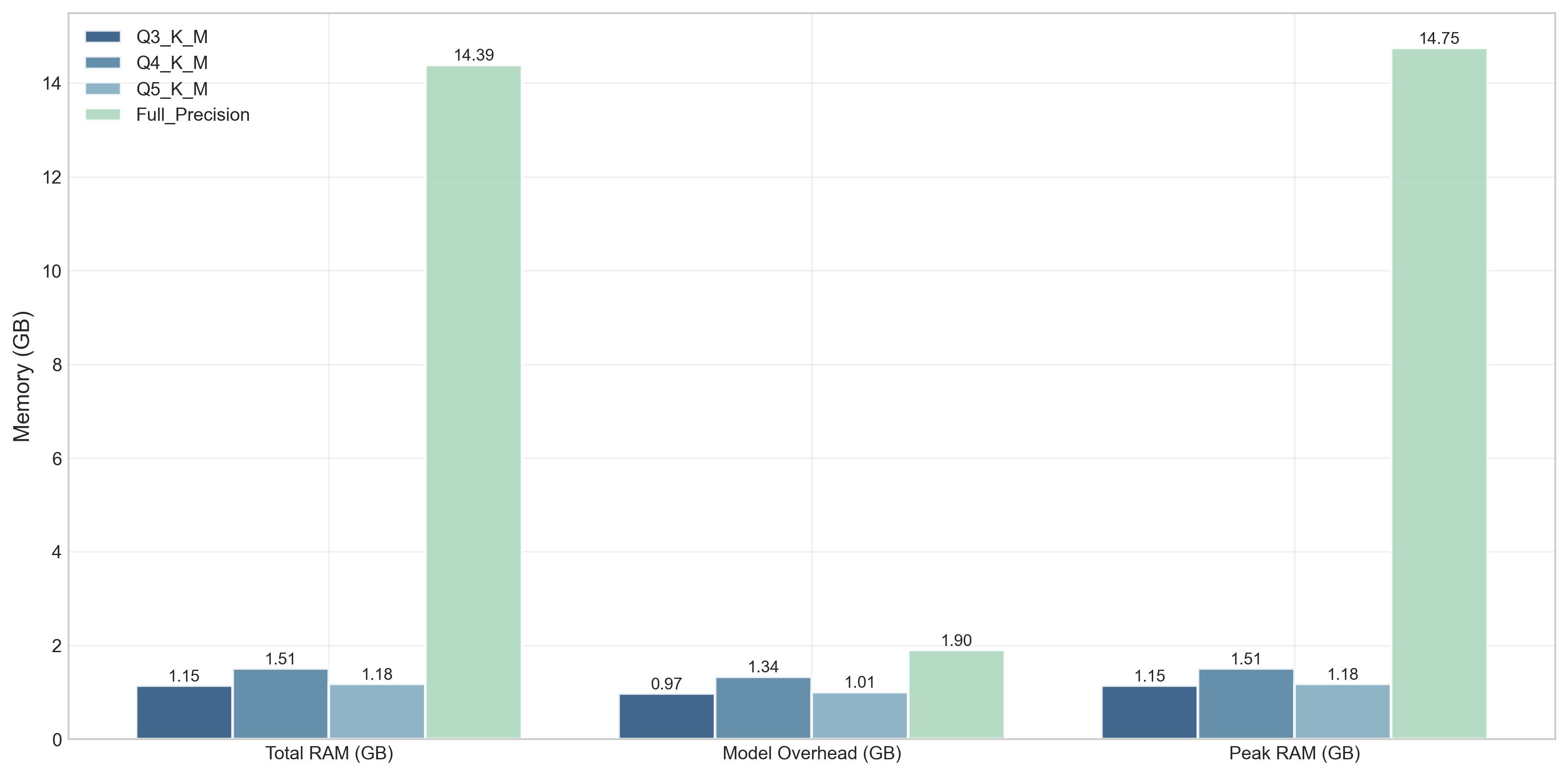}
  \caption{CPU Memory Usage Comparison. GGUF quantization reduces RAM consumption by over 90\%, from 14.39 GB to less than 1.6 GB.}
  \label{fig:cpu_memory}
\end{figure}

\begin{figure}[htbp]
  \centering
  \includegraphics[width=0.7\textwidth]{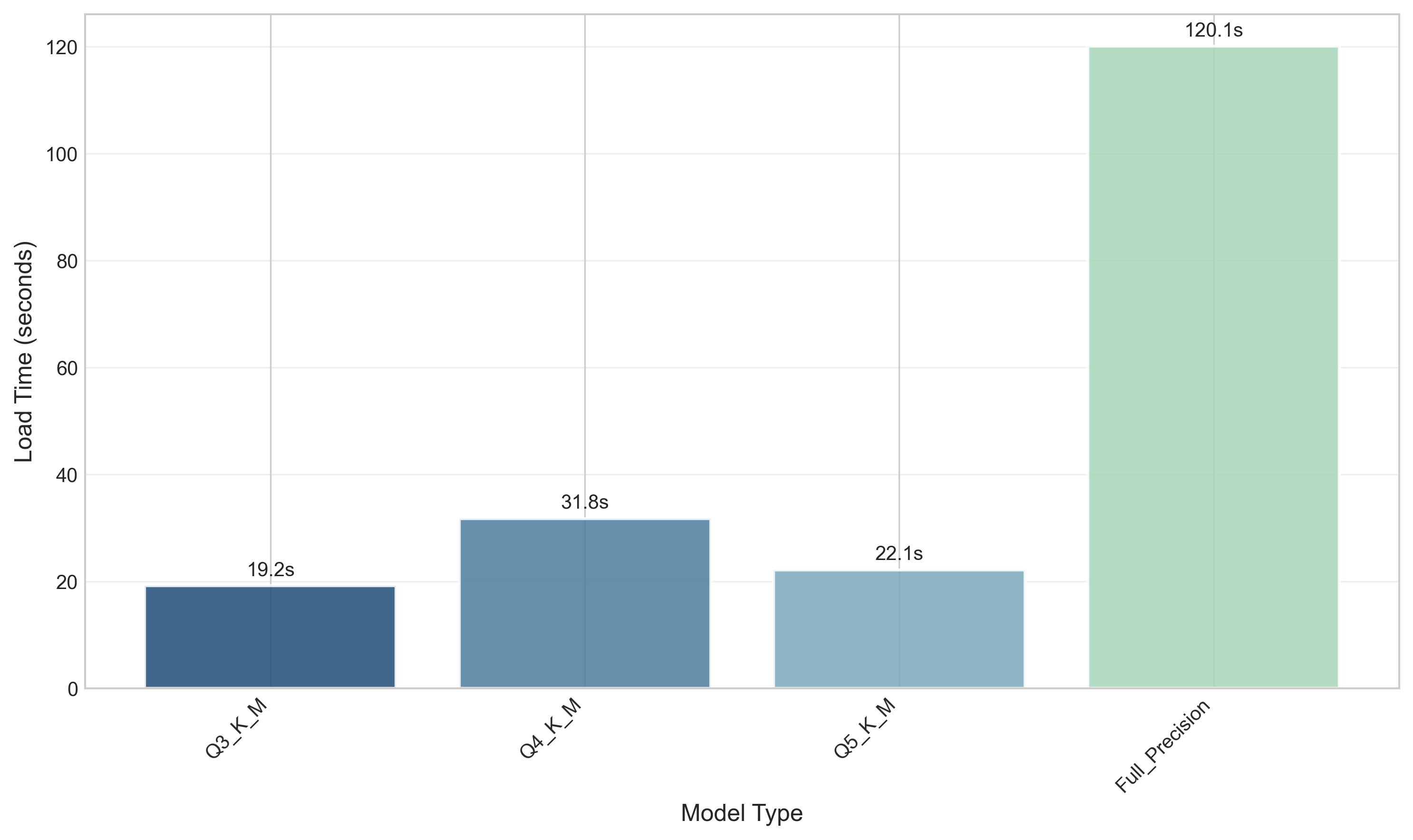}
  \caption{CPU Load Time Comparison. Quantized GGUF models load significantly faster than the full-precision version.}
  \label{fig:cpu_load_time}
\end{figure}

These speed advantages were complemented by dramatic reductions in RAM consumption and load times. Figure \ref{fig:cpu_memory} shows that the RAM footprint was reduced from 14.39 GB for the FP16 model to around 1.15-1.51 GB for the GGUF versions - a reduction of over 90\%. This makes it feasible to run the model on standard consumer hardware. Consequently, load times also saw a significant improvement, as shown in Figure \ref{fig:cpu_load_time}.

\subsection{Accuracy vs. Efficiency Trade-offs on CPU}
\begin{figure}[htbp]
  \centering
  \includegraphics[width=0.7\textwidth]{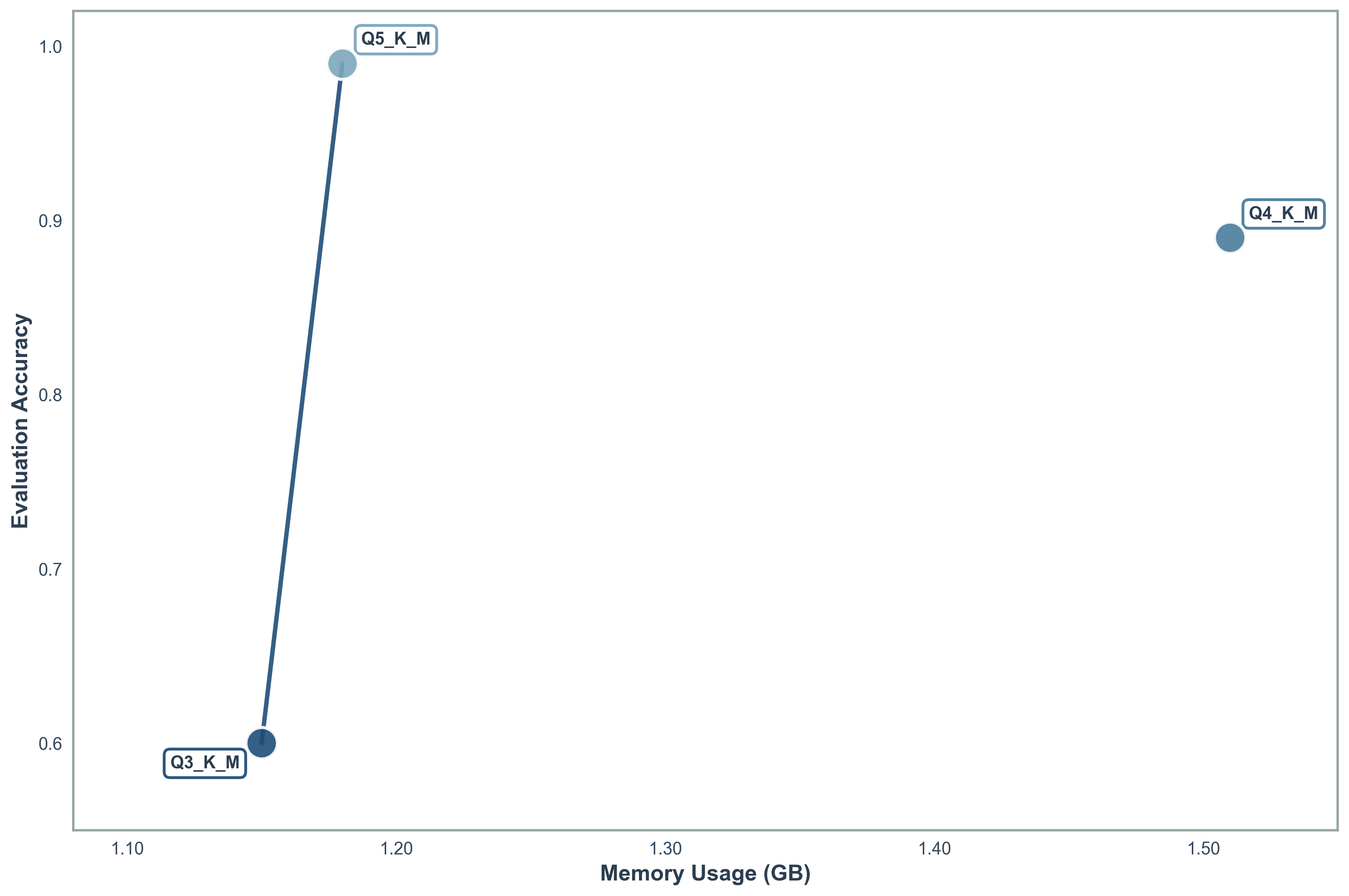}
  \caption{Pareto Frontier: Accuracy vs. Memory Usage (CPU). The Q5\_K\_M model offers the highest accuracy with minimal memory usage, while the Q3\_K\_M model is suboptimal due to its low accuracy.}
  \label{fig:pareto_memory}
\end{figure}

\begin{figure}[htbp]
  \centering
  \includegraphics[width=0.7\textwidth]{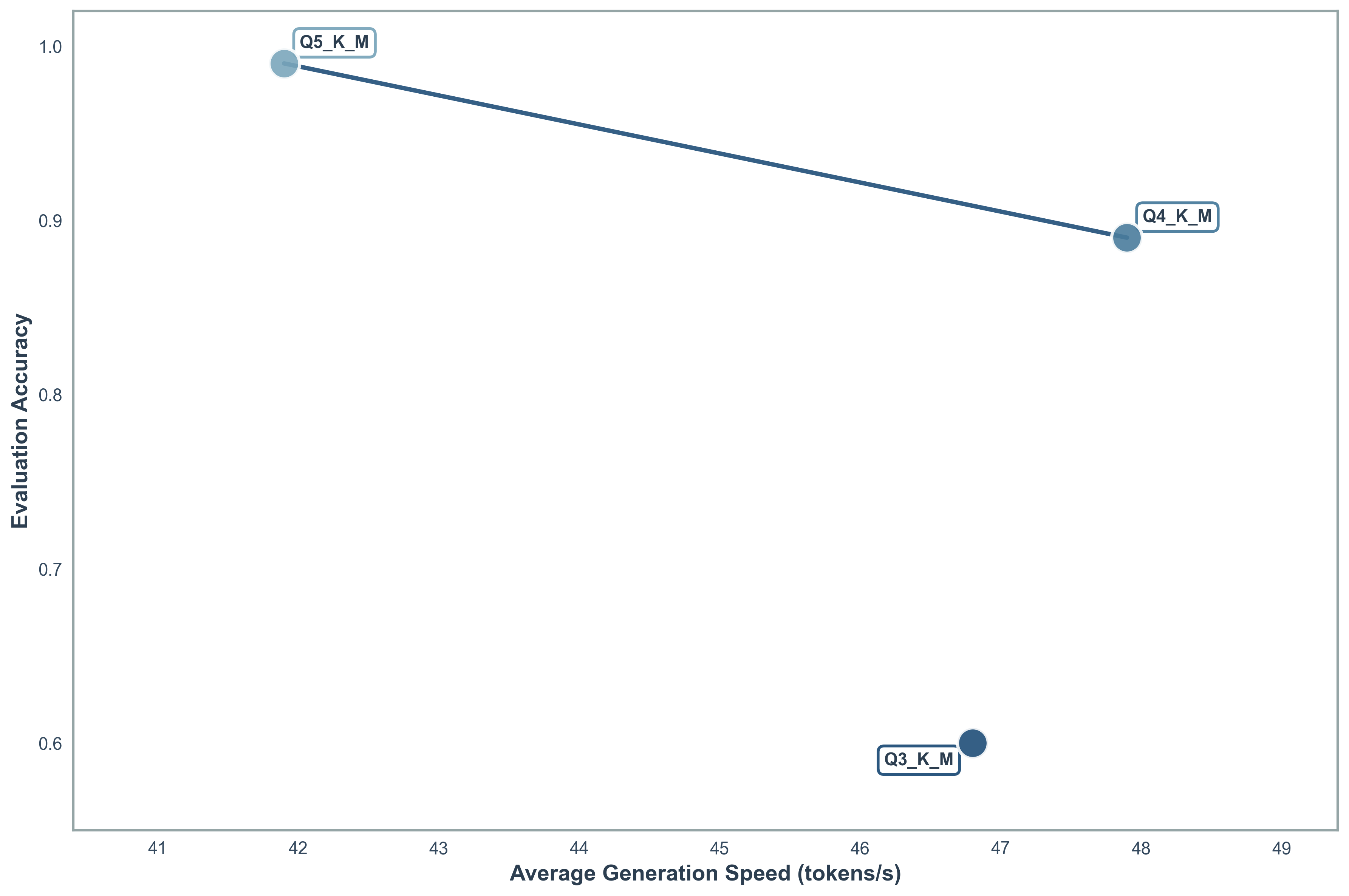}
  \caption{Pareto Frontier: Accuracy vs. Generation Speed (CPU). The Q4\_K\_M model provides the highest speed, while the Q5\_K\_M model offers the best accuracy, presenting a clear trade-off.}
  \label{fig:pareto_speed}
\end{figure}

The CPU performance results allow for a clear visualization of the trade-offs between accuracy, memory, and speed. Figures \ref{fig:pareto_memory} and \ref{fig:pareto_speed} plot the Pareto frontiers for the GGUF variants. These charts illustrate the optimal choices depending on application priorities.

The analysis reveals two optimal candidates for CPU deployment. The \textbf{Q5\_K\_M (5-bit)} model provides the highest possible accuracy (0.99) with excellent speed (around 42 tokens/s) and minimal memory usage. The \textbf{Q4\_K\_M (4-bit)} model offers the peak inference speed (nearly 48 tokens/s) at the cost of a moderate reduction in accuracy (0.89). The 3-bit version is clearly suboptimal, as it provides no significant memory or speed advantage over the other quantized versions but suffers a severe accuracy penalty.

\section{Discussion}

The results of our study offer a multi-faceted view of the opportunities and challenges associated with deploying small, optimized language models. Our findings confirm that specialization can elevate a small model to state-of-the-art performance; however, they also reveal that the efficiency gains from optimization techniques such as quantization are deeply intertwined with the underlying hardware and software ecosystem.

\subsection{Small Models Can Achieve State-of-the-Art Accuracy}
Our central finding - that a 1B parameter Llama 3.2 model can match the 99\% accuracy of GPT-4.1 after specialized fine-tuning - contributes to a growing body of evidence challenging the notion that larger models are constantly superior. This result underscores the power of domain specialization. While massive models possess a broad, generalist knowledge base, a smaller model trained on a high-quality, task-specific dataset can develop a deep, expert-level competency within its narrow domain. This principle has been demonstrated across diverse fields, from outperforming GPT-4 on arithmetic tasks \cite{liuGoatFinetunedLLaMA2023a} to achieving comparable performance in pedagogical applications \cite{solanoNarrowingGapSupervised2025} and medical language understanding \cite{fuBioMistralNLUMoreGeneralizable2025}. The comprehensive study in LoRA Land, which found that over 200 fine-tuned small models surpassed GPT-4 on specific tasks, solidifies this paradigm \cite{zhaoLoRALand3102024}. For many business applications, particularly in e-commerce where user interactions follow predictable patterns \cite{heroldLiLiuMEBaysLarge2024}, deploying a fleet of small, expert models is a more cost-effective, private, and computationally efficient strategy than relying on a single, oversized generalist model.

\subsection{Quantization is Not a Free Lunch: The Hardware-Software Synergy}
Perhaps the most critical insight from our performance analysis is that the benefits of quantization are not intrinsic to the algorithm itself but emerge from the synergy between the model format, the inference engine, and the hardware architecture. The stark contrast between our GPU and CPU results illustrates this point perfectly.

The 82\% slowdown of the 4-bit GPTQ model on the NVIDIA T4 GPU, despite a 41\% reduction in VRAM, highlights a common pitfall. The slowdown suggests that the T4 GPU did not execute computations directly in 4-bit. Instead, it appears that the quantized weights were converted back to higher precision during inference, adding overhead that negated the benefits of reduced memory bandwidth. This finding aligns with systematic characterizations of LLM quantization, which show that performance and energy gains are highly dependent on the interplay between the quantization scheme and the GPU's capabilities \cite{shiSystematicCharacterizationLLM2025, fernandezEnergyConsiderationsLarge2025}. On modern GPUs designed with native support for low-precision arithmetic, these results would likely be reversed, leading to significant speedups \cite{leeComprehensiveEvaluationQuantized2024}.

Conversely, the massive success of the GGUF formats on the CPU (achieving over 18× speedup) is a testament to software optimization. The llama.cpp library is purpose-built to leverage CPU vector instructions (such as AVX) for highly efficient low-bit integer matrix multiplications. This demonstrates that with the right software, even ubiquitous consumer hardware can become a powerful platform for LLM inference. This principle of hardware-software co-design is a central theme in efficient AI deployment, from large-scale data centers dynamically managing resources \cite{stojkovicDynamoLLMDesigningLLM} to hyper-efficient accelerators for edge devices \cite{tianCLONECustomizingLLMs, qiaoTeLLMeEnergyEfficientTernary2025}. Our results provide a clear, practical example of this principle in action.

\subsection{Practical Recommendations and the Pareto Frontier}
Our analysis of the accuracy-vs-efficiency trade-offs, particularly for CPU deployment (Figures \ref{fig:pareto_memory} and \ref{fig:pareto_speed}), allows us to formulate concrete recommendations for practitioners. The choice of which model variant to deploy is not a simple one but a strategic decision based on application-specific priorities.
\begin{itemize}
    \item \textbf{For Maximum Accuracy:} In scenarios where correctness is paramount, the 5-bit GGUF (Q5\_K\_M) model emerges as the optimal choice for CPU deployment. It delivers the full 99\% accuracy of the original fine-tuned model while still providing significant benefits in RAM reduction and a substantial inference speedup. For GPU deployment, using the FP16 model or a GPTQ-quantized version on modern hardware with native low-bit support would be the recommended path.
    \item \textbf{For Maximum Speed:} When lowest latency is the primary goal and a slight dip in accuracy is acceptable, the 4-bit GGUF (Q4\_K\_M) model is the superior option, offering the highest throughput on our CPU testbed.
    \item \textbf{Suboptimal Choices:} The 3-bit GGUF model is clearly a suboptimal choice. It is "dominated" on the Pareto frontier, offering no significant performance advantage over the 4-bit and 5-bit versions while suffering a catastrophic loss of accuracy. This finding aligns with broader studies showing that performance degradation at very low bit-depths can be non-linear and severe \cite{zhaoBenchmarkingPostTrainingQuantization2025}.
\end{itemize}
This decision-making process, balancing multiple objectives such as accuracy, latency, and energy, resonates with calls to move beyond single metrics and adopt more holistic evaluation frameworks, such as considering energy-per-token \cite{wilhelmTestTimeComputeStrategies2025, maliakelInvestigatingEnergyEfficiency2025}.

\subsection{Limitations of the Study}
While our findings provide valuable insights, it is important to acknowledge the limitations of this study, which also point to avenues for future research. First, our model was trained and evaluated exclusively on \textbf{synthetically generated data}. While carefully designed with strategic noise injection, synthetic data may not fully capture the complexity and unpredictability of real-world user queries. The challenges of ensuring realism and avoiding distributional mismatch in synthetic data are well-documented \cite{nadasSyntheticDataGeneration2025a}. Second, our \textbf{hardware scope} was limited to a single older-generation GPU (NVIDIA T4) and one type of consumer CPU. As discussed, performance results for GPTQ are expected to be significantly different and more favorable on newer GPUs. Third, our investigation focused on a \textbf{single, well-defined task}. The model's excellent performance in intent extraction does not guarantee similar success on other, more complex e-commerce tasks, such as those benchmarked in ShoppingBench, which require multi-step reasoning and tool use \cite{wangShoppingBenchRealWorldIntentGrounded2025}. Finally, our use of a strict \textbf{Exact Match Accuracy} metric, while appropriate for this task, is binary and does not capture nuances of partially correct answers.

\section{Conclusion}

This paper investigated the feasibility of using small, optimized open-weight language models as a practical alternative to large commercial systems for a specialized e-commerce task. Our findings demonstrate that this approach is not only viable but highly effective, offering a pathway to building state-of-the-art AI solutions that are both powerful and resource-efficient.

We have successfully shown that through parameter-efficient fine-tuning on a high-quality synthetic dataset, a 1 billion parameter Llama 3.2 model can achieve 99\% exact match accuracy in a structured intent recognition task. This result places its performance on par with the significantly larger, state-of-the-art GPT-4.1 model, confirming our primary hypothesis. However, our analysis of post-training quantization revealed a critical layer of complexity: the operational benefits are profoundly dependent on the deployment environment. On an older GPU architecture, 4-bit GPTQ quantization, while drastically reducing memory, led to a counter-intuitive 82\% slowdown in inference due to dequantization overhead. In stark contrast, GGUF formats on a CPU achieved over an 18× improvement in inference throughput and approximately 90\% lower RAM usage compared to the FP16 baseline, making sophisticated LLM inference feasible on consumer-grade hardware.

The primary conclusion of this work is that the future of many applied AI solutions may not lie solely with ever-larger generalist models, but in a diverse ecosystem of smaller, highly specialized, and hardware-aware models. For developers and organizations, our results provide a clear directive: the choice of an optimization strategy cannot be made in isolation from the target hardware. A model that is highly efficient in one context can be impractical in another. This paradigm enables organizations to develop customized solutions that are more cost-effective, private, and computationally efficient.

Building on these findings, several avenues for future research emerge. First, validating the fine-tuned model's performance on a large corpus of real-world, anonymized user data would provide the definitive confirmation of its practical utility. Second, replicating the GPU performance analysis on modern architectures (e.g., NVIDIA Ampere or Hopper) is essential to quantify the potential inference speedups that GPTQ can offer when paired with native low-precision hardware support. Finally, the scope of this work could be expanded by fine-tuning the model to handle a broader range of e-commerce tasks, such as product recommendation or order status inquiries, and by exploring alternative optimization techniques, such as AWQ quantization or other PEFT methods like DoRA.

Ultimately, this research serves as a practical demonstration that with the right specialization and optimization strategies, small models can indeed stand shoulder-to-shoulder with giants, offering a more accessible and sustainable path for the widespread adoption of advanced AI.

\printbibliography

\end{document}